\documentclass[10pt,twocolumn]{article}

\usepackage[utf8]{inputenc}
\usepackage{graphicx}
\usepackage{authblk}
\usepackage[top=0.75in, bottom=0.75in, left=0.7in, right=0.7in]{geometry}
\usepackage{setspace}
\usepackage{times}
\usepackage{titlesec}
\usepackage{hyperref}
\usepackage{enumitem}
\usepackage{url}
\usepackage{footnote}
\makesavenoteenv{tabular}
\usepackage{longtable}
\usepackage[table,xcdraw]{xcolor}
\usepackage{placeins}
\usepackage{booktabs}
\usepackage{float}
\usepackage{enumitem}
\usepackage{tabularx}
\usepackage{cuted}
\usepackage{caption}
\usepackage{subcaption} 
\usepackage{afterpage}

\hypersetup{
  pdftitle={IndianBailJudgments-1200},
  pdfauthor={Sneha Deshmukh, Prathmesh Kamble},
  colorlinks=true,
  linkcolor=blue,
  citecolor=blue,
  urlcolor=blue
}

\titleformat{\section}{\large\bfseries}{\thesection.}{0.5em}{}
\titleformat{\subsection}{\normalsize\bfseries}{\thesubsection.}{0.5em}{}

\title{\textbf{IndianBailJudgments-1200: A Multi-Attribute Legal NLP Dataset for Bail Order Understanding in India}}
\author[1]{Sneha Deshmukh}
\author[1]{Prathmesh Kamble}
\affil[1]{Department of Computer Engineering, Datta Meghe College of Engineering\\
\texttt{sneha.deshmukh@dmce.ac.in} \hspace{2em}
\texttt{prathmesh.kamble@dmce.ac.in}}
\date{}

\begin{document}
\maketitle

\begin{abstract}
Legal Natural Language Processing (NLP) remains a low-resource field in jurisdictions like India, where access to high-quality, structured legal data is limited. This paper presents \textit{IndianBailJudgments}, a novel dataset comprising 1200 Indian court judgments related to bail decisions. Each case is annotated with over 20 structured attributes, including bail outcome, IPC sections, crime type, court name, and legal reasoning. These annotations were generated using a prompt-engineered GPT-4o (released May 2024) and manually verified for a subset of cases to ensure contextual reliability.

Our dataset supports diverse NLP tasks including case outcome classification, information extraction, legal summarization, and fairness analysis. It is the first public dataset focused solely on Indian bail jurisprudence. We aim to link real-world legal texts with AI-driven analysis and offer this resource openly for legal NLP research.
\end{abstract}

\section{Introduction}

The intersection of legal technology and artificial intelligence has rapidly evolved in recent years, with Natural Language Processing (NLP) playing a central role in enabling machines to understand, summarize, and analyze complex legal texts. However, the majority of publicly available legal datasets originate from Western jurisdictions such as the U.S. or Europe, leaving the Global South—particularly India—underrepresented in the landscape of legal NLP resources.

India, with its vast and multilingual judicial system, generates thousands of judgments each year. A single bail order may weigh multiple factors—crime severity, prior record, and social context—often hidden in lengthy legal prose. Yet access to structured, high-quality legal data remains severely limited. This paper introduces \textit{IndianBailJudgments}, a new benchmark dataset of 1200 Indian court decisions specifically related to bail—an understudied but socially critical aspect of the legal process.

\subsection{The Importance of Bail Decisions in India}

Bail orders are pivotal legal determinations that directly impact individual liberty and pre-trial detention. In India, over 75\% of the jail population consists of undertrial prisoners,~\cite{ncrb2022} contributing significantly to prison overcrowding. Securing bail is often challenging due to procedural delays, judicial discretion, limited legal representation, and socio-economic disparities. A well-reasoned bail judgment weighs multiple factors, including crime severity, prior record, co-accused parity, and gender of the accused.

Despite the critical role of bail jurisprudence, there exists no large-scale, structured dataset to study these decisions in the Indian context. Understanding patterns in bail outcomes is essential not only for legal research, but also for enabling fair policy reforms and access to justice. This dataset addresses that gap through multi-attribute annotations on 1200 real-world bail cases, capturing court, crime, outcome, and judicial reasoning dimensions.

\subsection{Why Legal NLP is Hard in India}

Developing NLP systems for Indian legal data presents several unique challenges:

\begin{itemize}[leftmargin=1.5em, itemsep=3pt, topsep=3pt]
    \item \textbf{Unstructured Sources:} Most Indian judgments exist as PDFs or scraped HTML, lacking standardized formatting or machine-readable metadata.
    
    \item \textbf{Multilingual and Inconsistent Style:} Documents often mix English with regional phrasing, and formatting varies significantly across courts.
    
    \item \textbf{Sparse Annotations:} Unlike U.S. or commercial datasets, Indian legal texts rarely come with labeled outcomes or structured judicial rationale.

    \item \textbf{No Open Benchmarks:} There is a shortage of public Indian legal datasets with detailed annotations for NLP tasks like classification or summarization.
\end{itemize}

As a result, existing legal NLP models struggle to adapt to Indian court language. Our dataset addresses these gaps through prompt-based LLM annotation, schema design guided by legal experts, and structured JSON releases for reproducible legal AI research.

\begin{table*}[t]
\centering
\small
\renewcommand{\arraystretch}{1.25}
\begin{tabular}{@{}p{3.5cm} p{4cm} p{2cm} p{5.5cm}@{}}
\toprule
\textbf{Dataset} & \textbf{Jurisdiction} & \textbf{Size} & \textbf{Tasks Enabled} \\
\midrule
ECtHR Cases~\cite{chalkidis2019ecthr} & European Court of Human Rights & 11K & Judgment prediction, label classification \\
CUAD~\cite{hendrycks2021cuad} & U.S. Commercial Contracts & 13K clauses & Contract clause extraction, NER \\
LEXGLUE~\cite{chalkidis2021lexglue} & Europe/U.K./U.S. & Varies & Multi-task: summarization, classification, entailment \\
INDIANLEGAL-BERT~\cite{jain2021indianlegalbert} & India & 8M docs & Pretraining for Indian legal NLP \\
ILDC~\cite{malik2021ildc} & Indian Supreme/High Court & 35K & Summarization \\
BillSum~\cite{korzen2020billsum} & U.S. Congressional Bills & 23K & Summarization \\
SCOTUS~\cite{caselawaccessproject} & U.S. Supreme Court & 50K+ & Opinion classification \\
\bottomrule
\end{tabular}
\caption{Comparison of major legal NLP datasets across jurisdictions.}
\label{tab:dataset-comparison}
\end{table*}

\section{Related Work}

\subsection{Legal NLP Datasets Across Jurisdictions}

The legal NLP community has developed several benchmark datasets across jurisdictions, enabling tasks such as contract understanding, court judgment prediction, statute retrieval, and interpretation of law. These datasets span multiple legal systems, including common law, civil law, and mixed jurisdictions, allowing researchers to evaluate models on diverse linguistic and structural patterns in legal texts.

Many datasets are annotated with legal-specific labels—such as case outcomes, citations, and charges—providing rich supervision for training machine learning models. Cross-jurisdictional resources also enable comparative legal research and foster transfer learning approaches that generalize across systems.

Despite these advances, few datasets offer the granular, multi-attribute annotations necessary to explore the nuanced reasoning processes behind judicial decisions.

Table~\ref{tab:dataset-comparison} summarizes key characteristics of prominent datasets, including their jurisdiction, task type, annotation style, and size. This growing ecosystem is accelerating progress in legal AI and bridging the gap between law and technology.

\subsection{Gaps in Indian Legal Datasets}

While recent efforts like INDIANLEGAL-BERT and ILDC have made significant strides in collecting Indian legal documents, most datasets remain limited to raw text or sentence-level labels. They lack labeled attributes that capture legal reasoning, bail status, gender, or crime type—fields essential for tasks like fairness audits, outcome prediction, and jurisprudence analysis.

Moreover, Indian legal datasets often suffer from inconsistent formatting, lack of metadata, and absence of gold-standard annotations validated by legal experts. Our dataset addresses these gaps by providing multi-attribute structured annotations on real-world court judgments, along with verified labels and task-ready formats. It is the first of its kind to focus specifically on Indian bail jurisprudence.

\section{Dataset Creation}

\subsection{Data Collection Sources}

A dataset of 1200 bail-related court judgments was curated from publicly available Indian legal repositories, primarily Indian Kanoon~\cite{indiankanoon}, which aggregates decisions from various High Courts across India. The goal was to build a representative, diverse sample of Indian bail jurisprudence across time, jurisdiction, and crime type.

To ensure meaningful variation, we carefully selected cases spanning both fresh bail applications and bail cancellation scenarios. High Courts were prioritized due to their role in setting precedent and addressing complex bail petitions. We ensured geographic coverage across northern, western, southern, and eastern zones, including courts like the Bombay High Court, Allahabad High Court, and Madras High Court.

In pursuit of diversity, the dataset spans a wide range of crime categories such as murder, narcotics offenses, sexual offenses, dowry harassment, and cybercrime. The sampling strategy was calibrated to avoid overrepresentation of any single offense category or region. Case metadata was manually tracked during the scraping process to ensure diversity across the gender of the accused, type of bail sought, and judicial outcomes (granted/rejected). Documents were collected in either plain text or PDF formats, then preprocessed to remove noise such as repeated headers, HTML artifacts, and procedural boilerplate.

In addition to maximizing diversity, we also ensured temporal variation by including judgments from different years to account for shifts in judicial rationale and policy interpretations over time. This temporal spread enables researchers to examine longitudinal patterns, such as how courts’ attitudes toward specific offenses or social factors may have evolved. Moreover, the inclusion of both bail approvals and denials provides a balanced foundation for training outcome prediction models without inherent label skew.

To promote reproducibility, we maintained logs of case PDFs, court metadata, and processing status at every stage of pipeline development. This ensures transparency in data provenance and allows future researchers to extend or audit the dataset construction process.

\begin{figure*}[t]
    \centering
    \includegraphics[width=\textwidth]{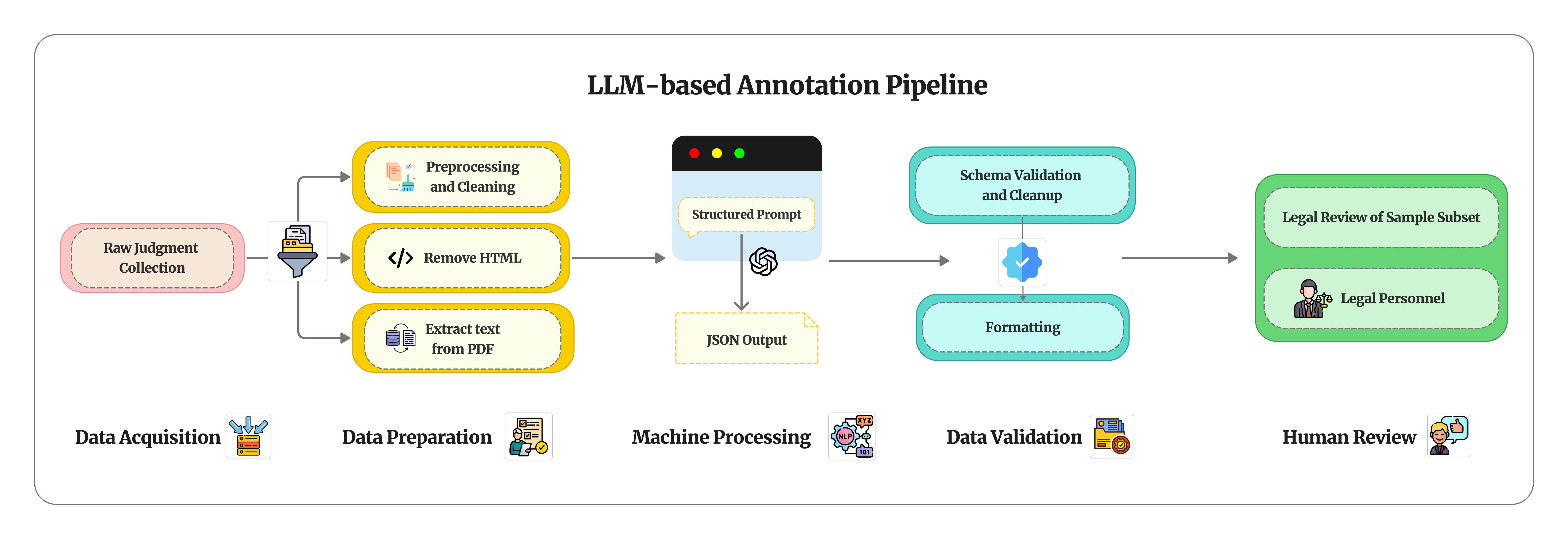} 
    \caption{Overview of the LLM-based annotation pipeline used to structure Indian bail judgments.}
    \label{fig:annotation-flow}
\end{figure*}

\subsection{Schema Design Philosophy}

A schema with over 20 structured attributes per case was designed to support a wide range of legal NLP tasks and downstream applications. These attributes were selected to reflect both factual aspects (e.g., crime type, IPC sections, prior cases) and legal reasoning (e.g., legal issues raised, summary, judgment reason).

The schema was inspired by real-world legal reasoning processes and iteratively refined through consultations with legal personnel. It includes:
\begin{itemize}
    \item \textbf{Binary and categorical fields} such as \texttt{bail\_outcome}, \texttt{bail\_type}, \texttt{prior\_cases}, and \texttt{bias\_flag}, which support supervised learning and fairness analysis.
    \item \textbf{Free-text fields} like \texttt{facts}, \texttt{legal\_issues}, and \texttt{judgment\_reason}, which enable summarization, entailment, and retrieval tasks.
    \item \textbf{Structured lists} such as \texttt{ipc\_sections} and \texttt{legal\_principles\_discussed}, which are useful for sequence labeling and span extraction tasks.
\end{itemize}

Importantly, the schema was designed to accommodate two types of cases: individual bail decisions (Type 1) and landmark principle-setting rulings (Type 2). Placeholder values such as \texttt{"Unknown"} or \texttt{"Not specified"} were used to enforce consistency in edge cases and missing values.

Overall, the schema balances legal relevance with machine learning usability, making it suitable for a variety of applications ranging from classification to explainable AI research. The annotated dataset, described in the next section, was generated in line with this schema.

\section{Annotation via LLMs}

\subsection{Prompt-Driven Annotation Pipeline}

An annotation pipeline powered by OpenAI’s GPT-4o model was developed to transform raw, unstructured court judgments into structured, research-grade data. A carefully engineered prompt was designed to extract more than 20 fields per case, capturing legal, procedural, and factual dimensions of bail decisions.

The prompt was built to handle two primary case types:
\begin{itemize}
    \item \textbf{Type 1:} Individual bail applications with a final outcome (granted or rejected).
    \item \textbf{Type 2:} Landmark or precedent-setting cases that interpret legal doctrines.
\end{itemize}

Output formatting was strictly enforced. All boolean values (e.g., \texttt{bias\_flag}, \texttt{bail\_cancellation\_case}) were lowercase. Fields like \texttt{ipc\_sections} and \texttt{legal\_principles\_discussed} returned JSON-style lists, and missing information was encoded using placeholders such as \texttt{"Unknown"} or \texttt{"Not specified"}.

\subsection{Preprocessing and Execution}

Prior to annotation, each judgment was passed through an OCR and cleanup pipeline to extract meaningful text from scanned PDFs or HTML-heavy court sources. This step involved removing line breaks, duplicative headers, and procedural noise often found in legal documents.

Once cleaned, the judgment was fed into the LLM along with the prompt. The model produced structured outputs in JSON format, adhering to our schema  (see Figure~\ref{fig:annotation-flow}). A post-processing validation script checked for formatting consistency, required field presence, and schema alignment.

\subsection{Legal Review and Verification}

To ensure contextual reliability, a subset of 150 cases (approximately 12.5\% of the corpus) was manually reviewed by multiple individuals with formal legal training. These legal personnel were briefed on the annotation schema and evaluated the LLM-generated outputs for correctness, coherence, and interpretability. The reviewers found the annotations largely accurate and well-aligned with how Indian courts frame bail decisions, with only minor inconsistencies observed in edge cases such as overlapping bail types.

While this legal review does not substitute for judicial or certified legal annotation, it affirms the dataset's quality for downstream machine learning tasks. It also serves as a proof-of-concept for combining large language models with domain-informed schema design in low-resource domains.

\subsection{Scalability and Future Directions}

This annotation pipeline exemplifies a scalable, replicable methodology for creating high-quality legal datasets without requiring extensive manual labor. It leverages the rapid reasoning capability of LLMs while incorporating human-in-the-loop feedback for improved reliability.

Future iterations may incorporate automatic field-level confidence scoring, hybrid rule+LLM models for edge cases, and task-specific prompt tuning to enhance interpretability and robustness.

\begin{figure}[ht]
    \centering
    \includegraphics[width=0.95\linewidth]{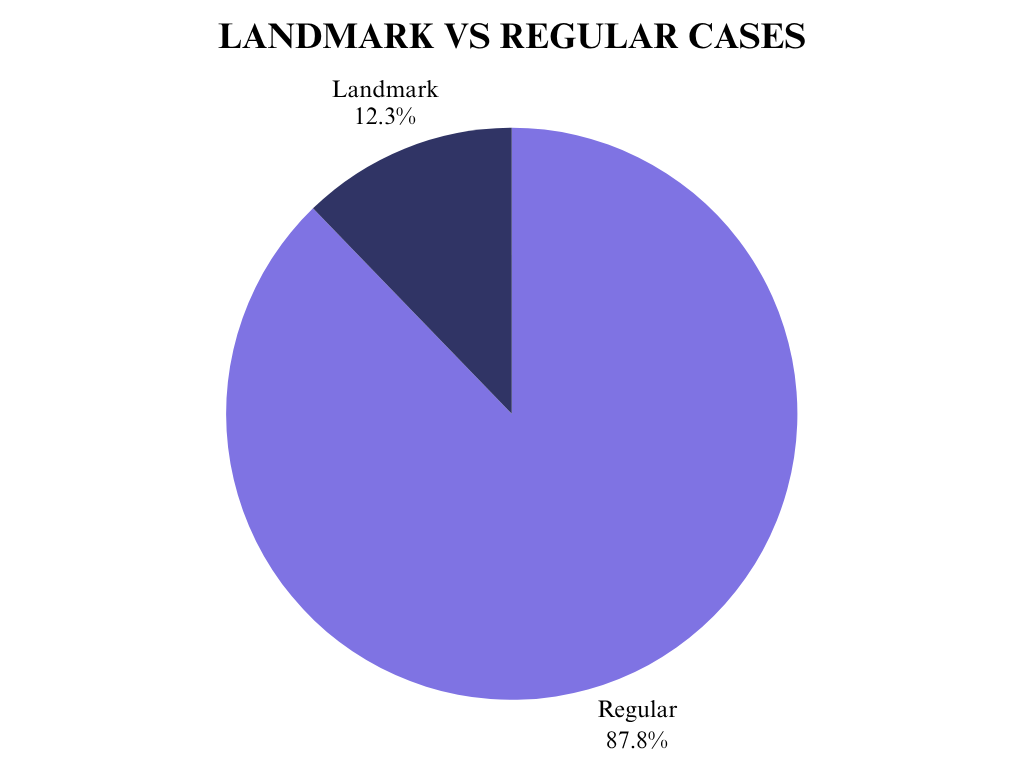}
    \caption{Landmark vs regular cases}
    \label{fig:landmark_vs_regular}
\end{figure}

\section{Dataset Statistics}

The \textit{IndianBailJudgments} dataset comprises 1200 bail-related court judgments, each annotated with over 20 structured fields. The dataset spans multiple High Courts in India and includes a diverse range of crime types, legal outcomes, and judicial reasoning patterns. It covers both offenses under the Indian Penal Code (IPC) and Special Acts such as NDPS and POCSO, offering a representative snapshot of India’s bail jurisprudence across jurisdictions and legal complexities.

The following statistical summaries, generated via a custom Python script, showcase the distribution of key attributes such as accused gender, crime type, bail outcome (granted or rejected), and bail type (regular, anticipatory, interim). Additional fields like the presence of legal arguments (e.g., parity or bias), landmark status, and court origin add analytical depth. These charts help visualize the dataset’s internal diversity and readiness for tasks like classification, fairness analysis, or case outcome prediction.

Beyond simple frequency analysis, the dataset supports exploratory correlation studies between attributes. For example, researchers can examine how bail decisions vary by gender within specific crime types or assess whether certain courts show higher rejection tendencies. Parity arguments can also be analyzed in relation to outcome success. Together with the structured schema (see Table~\ref{tab:schema}), these insights make the dataset a powerful foundation for empirical legal research, fairness audits, and interpretable legal AI.

\begin{figure}[ht]
    \centering
    \includegraphics[width=0.95\linewidth]{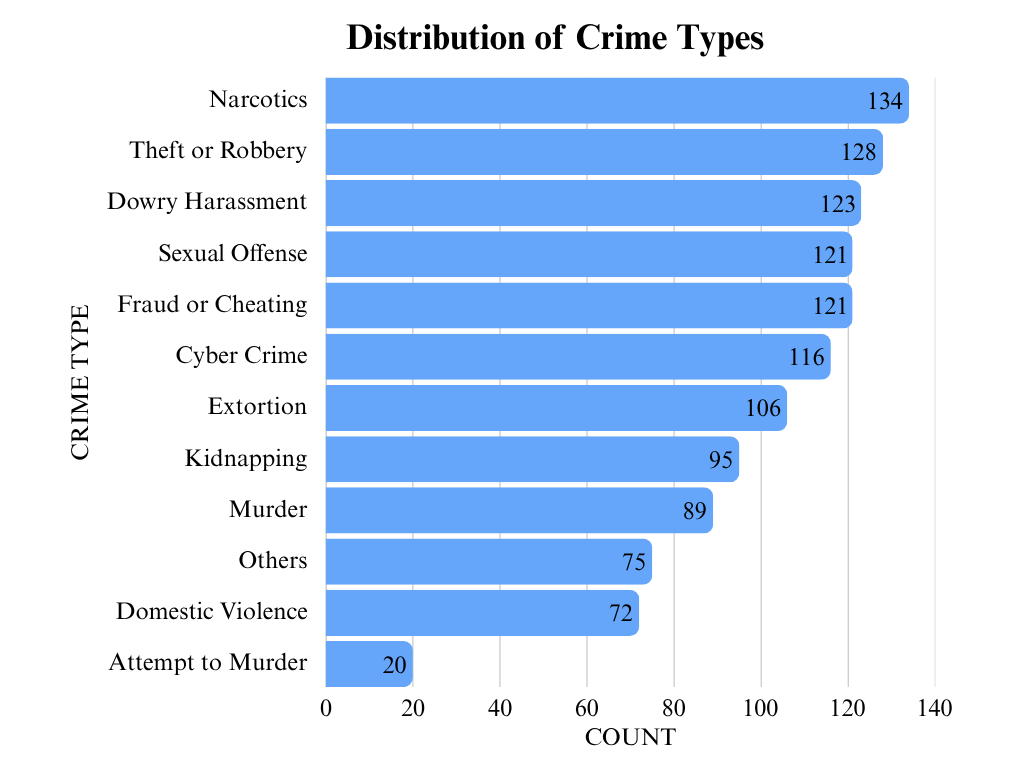}
    \caption{Distribution of crime types in the dataset}
    \label{fig:crime-types}
\end{figure}

\begin{figure}[ht]
    \centering
    \includegraphics[width=0.95\linewidth]{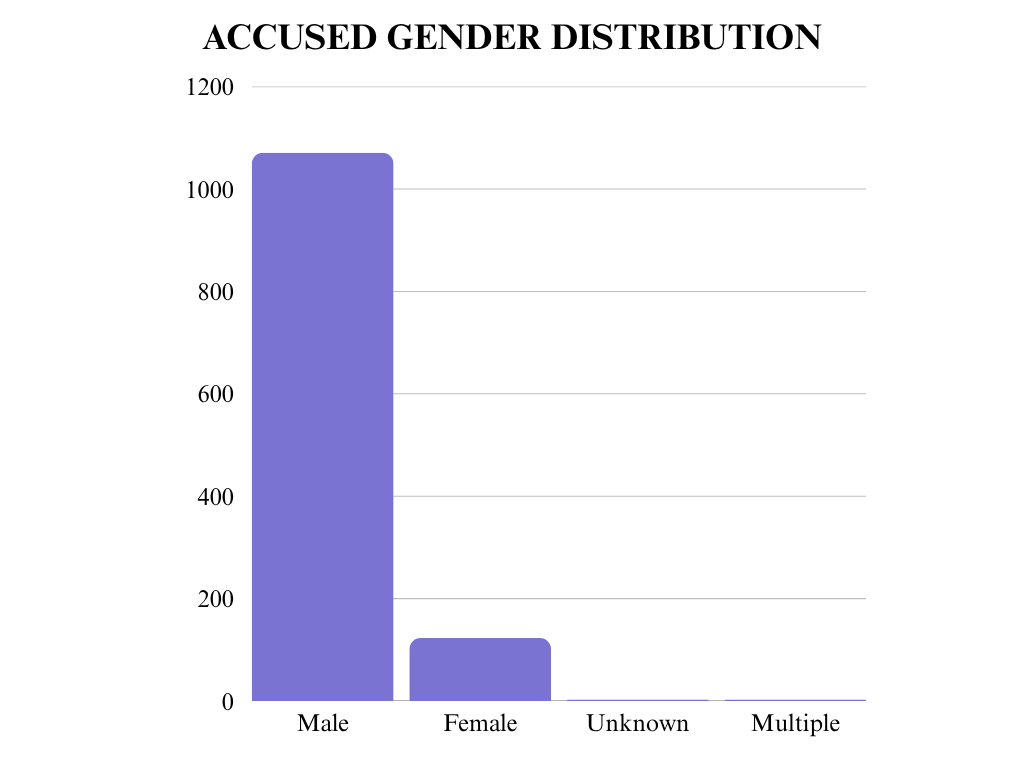}
    \caption{Accused Gender}
    \label{fig:gender_distribution}
\end{figure}

Figure~\ref{fig:crime-types} illustrates that the dataset encompasses a broad array of criminal charges, with offenses like murder, narcotics, and dowry harassment being among the most prevalent. This diversity reflects the heterogeneity of cases typically brought before Indian courts, enabling researchers to study bail decisions across a spectrum of legal and factual contexts. The inclusion of both severe and non-severe charges ensures that models trained on this dataset are not skewed toward a particular subset of criminal jurisprudence.

Gender distribution reveals a significantly male-dominated accused population, as shown in Figure~\ref{fig:gender_distribution}, a trend consistent with broader patterns in justice systems globally. This imbalance not only mirrors real-world data but also underscores the importance of gender-aware fairness research. For instance, certain bail considerations—such as caregiving responsibilities, risk assessment, or societal perceptions of dangerousness—may play out differently based on gender. Without capturing and analyzing these dynamics, AI systems risk reinforcing existing biases under the guise of objectivity.

This skew in gender representation may also influence judicial reasoning in subtle ways, particularly in cases involving vulnerable or marginalized populations. By making demographic and legal attributes explicit, the dataset enables researchers to identify patterns of differential treatment and construct fairness-aware machine learning models. It supports the development of bias-detection tools and explainable AI systems that can flag potential disparities in judicial outcomes.

Additionally, the dataset’s richness in legal argument types, demographic indicators, and outcome variables makes it a valuable resource not just for NLP and AI communities, but also for sociologists, criminologists, and policy researchers seeking to understand systemic patterns in Indian bail jurisprudence.

\begin{figure}[ht]
    \centering
    \includegraphics[width=0.95\linewidth]{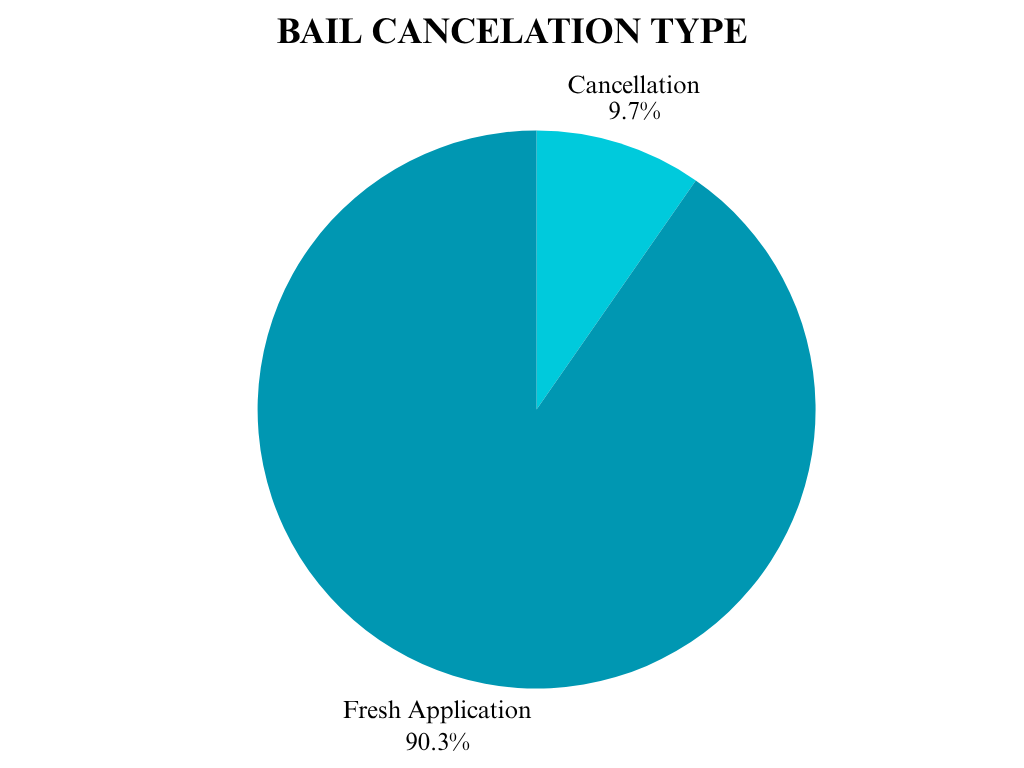}
    \caption{Bail cancellation types}
    \label{fig:bail_cancellation_type}
\end{figure}

\begin{figure}[ht]
    \centering
    \includegraphics[width=0.95\linewidth]{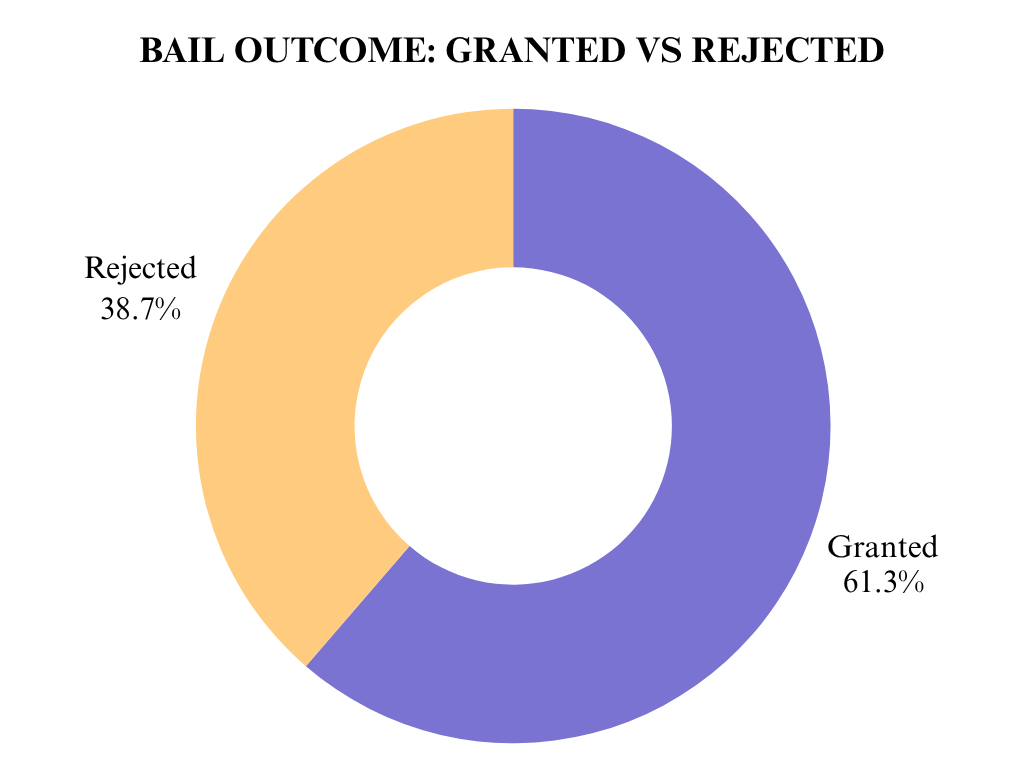}
    \caption{Bail outcomes}
    \label{fig:bail_outcome_distribution}
\end{figure}

Roughly 10\% of the cases in the dataset involve bail cancellation proceedings as shown in Figure~\ref{fig:bail_cancellation_type} , while the remaining 90\% are fresh bail applications. This distinction introduces valuable complexity into the dataset, enabling the development of models that are sensitive to both initial and review-stage judicial decisions. Cancellation cases often reflect nuanced court’s logic, such as violations of bail conditions or new evidence surfacing post-release—scenarios that can support multi-phase decision modeling or time-aware legal analytics.

Additionally, the distribution of granted versus rejected outcomes is relatively balanced which is evident in Figure~\ref{fig:bail_outcome_distribution}, minimizing risks of class imbalance and label skew that can hinder classification performance. This makes the dataset well-suited for training robust and generalizable supervised models, and facilitates fairer evaluation across metrics such as precision, recall, and F1-score. The balanced nature of these outcomes also offers a compelling opportunity for research in causal inference and counterfactual analysis: for example, investigating how specific combinations of factors—such as IPC sections invoked, prior criminal history, or gender—affect the likelihood of bail being granted or denied.

Furthermore, the presence of cancellation cases introduces scope for sequential learning frameworks, where models are trained to take prior decisions into account when predicting legal outcomes. These aspects make the dataset particularly rich for studying procedural dependencies and judicial consistency across time and courts.

\begin{figure}[ht]
    \centering
    \includegraphics[width=0.95\linewidth]{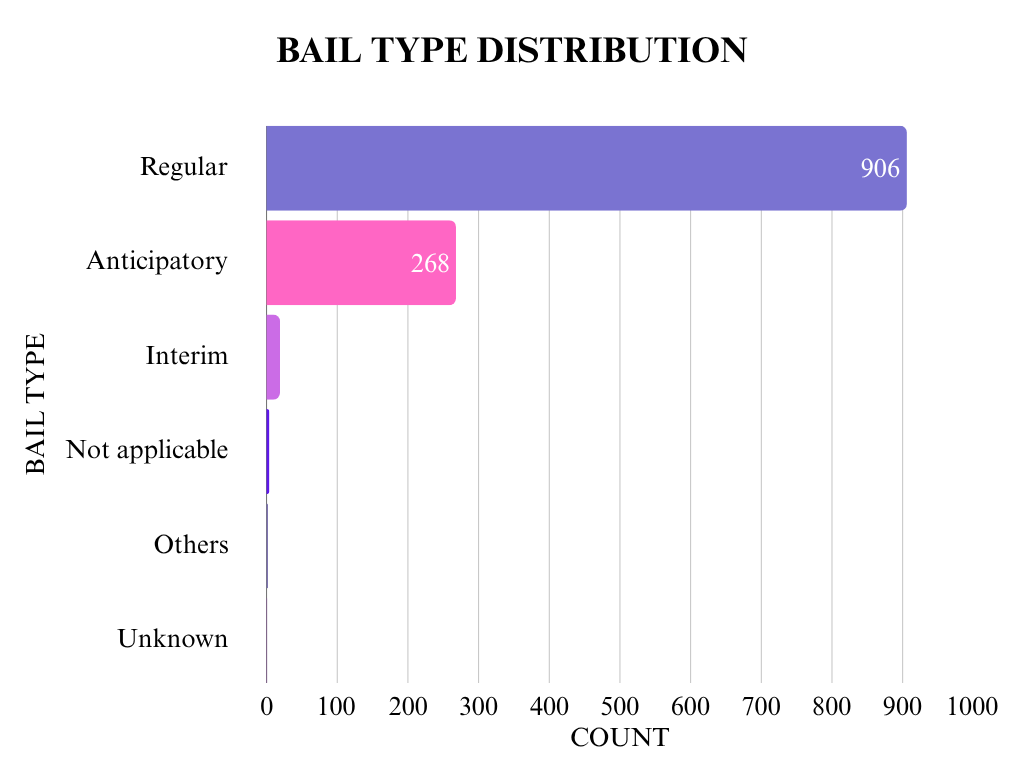}
    \caption{Bail types}
    \label{fig:bail_type_distribution}
\end{figure}

Regular bail dominates the dataset, while anticipatory and interim bails are underrepresented—highlighting the need for targeted data augmentation and improved balance as seen in Figure ~\ref{fig:bail_type_distribution}. This distinction is especially important for models that must reason across varied legal contexts, as each bail type has distinct procedural and jurisprudential implications. A better distribution would improve completeness and enable specialized models for currently underexplored bail types. Future versions may incorporate additional judgments or targeted scraping efforts to enhance this critical coverage.

\begin{table*}[ht]
\centering
\small
\begin{tabular}{@{}p{6cm} p{4cm} p{6cm}@{}}
\toprule
\textbf{Field} & \textbf{Type} & \textbf{Description} \\
\midrule
\texttt{case\_id} & String & Unique identifier \\
\texttt{case\_title} & String & Full case title \\
\texttt{court} & String & Name of court \\
\texttt{date} & Date & Judgment date \\
\texttt{judge} & String & Judge(s) name \\
\texttt{ipc\_sections} & List[String] & IPC/NDPS sections \\
\texttt{bail\_type} & String & Regular, Anticipatory, or Interim \\
\texttt{bail\_cancellation\_case} & Boolean & Bail review or cancellation flag \\
\texttt{landmark\_case} & Boolean & Principle-setting case \\
\texttt{accused\_name} & String & Name or \texttt{"Not specified"} \\
\texttt{accused\_gender} & String & \texttt{"Male"}, \texttt{"Female"}, or \texttt{"Unknown"} \\
\texttt{prior\_cases} & String & \texttt{"Yes"}, \texttt{"No"}, \texttt{"Unknown"} \\
\texttt{bail\_outcome} & String & \texttt{"Granted"}, \texttt{"Rejected"} \\
\texttt{bail\_outcome\_label\_detailed} & String & Outcome description \\
\texttt{crime\_type} & String & Crime category \\
\texttt{facts} & String & Short case summary \\
\texttt{legal\_issues} & String & Main legal questions \\
\texttt{judgment\_reason} & String & Court’s legal reasoning \\
\texttt{summary} & String & 2-line case summary \\
\texttt{bias\_flag} & Boolean & \texttt{true} if bias flagged \\
\texttt{parity\_argument\_used} & Boolean & \texttt{true} if co-accused parity applied \\
\texttt{legal\_principles\_discussed} & List[String] & Key doctrines discussed \\
\texttt{region} & String & State or jurisdiction \\
\bottomrule
\end{tabular}
\caption{Dataset Schema with Field Types and Descriptions.}
\label{tab:schema}
\end{table*}

\FloatBarrier

\begin{figure}[ht]
    \centering
    \includegraphics[width=0.95\linewidth]{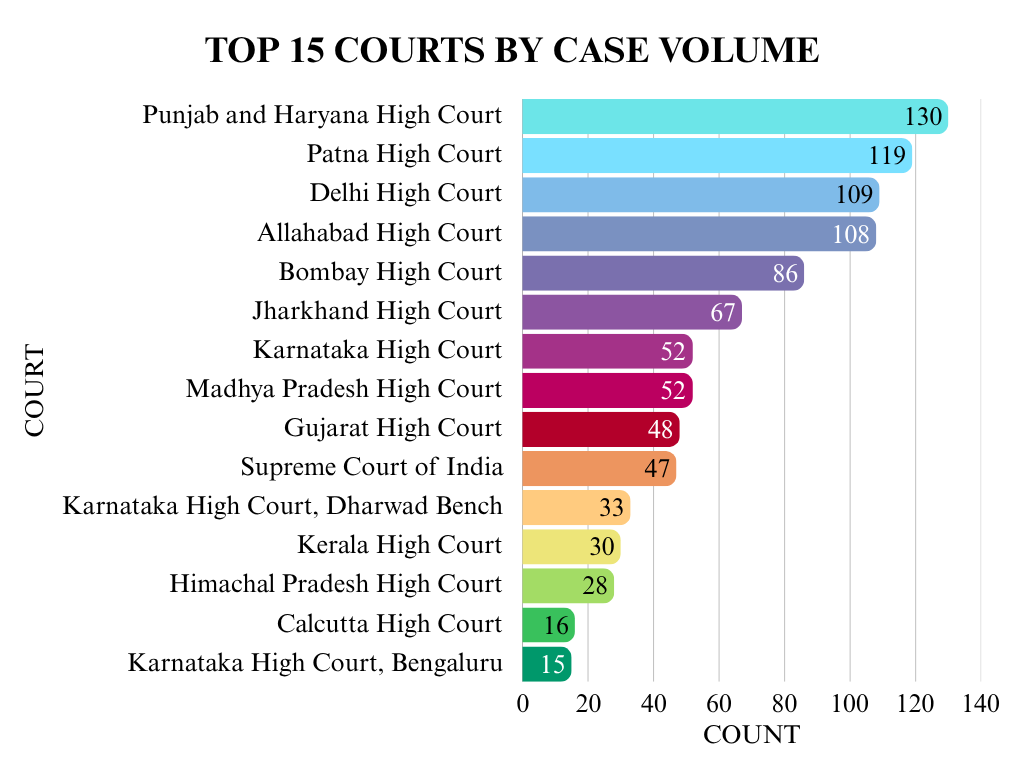}
    \caption{Top 15 courts}
    \label{fig:top_15_courts}
\end{figure}

These figures highlight the jurisdictional spread and statutory diversity in our dataset as illustrated in Figure ~\ref{fig:top_15_courts}. The inclusion of bail cases from High Courts such as Bombay, Allahabad, and Delhi ensures a well-rounded regional representation, capturing variations in judicial reasoning across different states.

\begin{figure}[ht]
    \centering
    \includegraphics[width=0.95\linewidth]{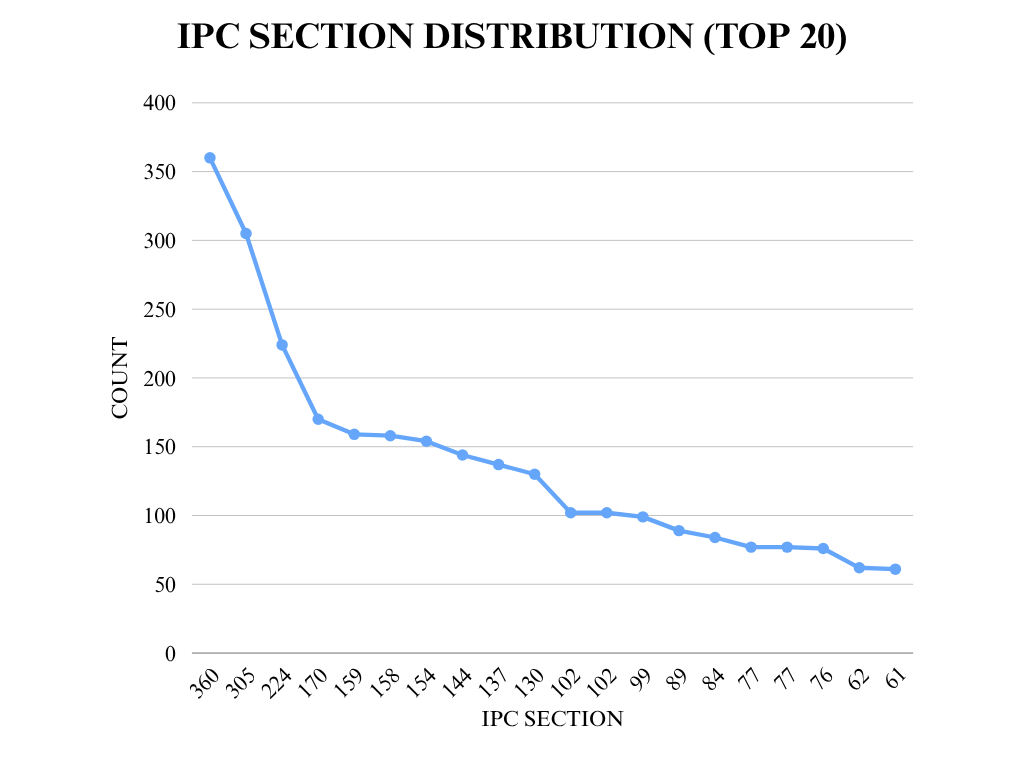}
    \caption{Top 20 IPC sections}
    \label{fig:ipc_section_distribution}
\end{figure}

The most common IPC sections in the dataset include 360 (kidnapping), 305 (abetment of suicide), 224 (resistance to lawful apprehension), and 170 (impersonating a public servant), followed by a range of sections such as 159, 154, 144, and 137 as can be seen in Figure~\ref{fig:ipc_section_distribution}. This statutory diversity spans both serious and procedural offenses, offering a granular lens into the legal basis on which bail is argued and granted or denied.

Such a wide distribution enables researchers to analyze charge-specific bail trends, study the prevalence of repeat charges, and explore correlations between specific statutes and judicial outcomes. It also enhances the dataset’s utility for statutory classification tasks and builds a foundation for developing AI models that can reason about legal provisions in the context of bail jurisprudence.

\FloatBarrier

In addition to the above, our dataset captures nuanced legal and social features such as whether co-accused parity was argued by the defense, whether judicial bias (e.g., gender, caste) was potentially present, and whether the case was recognized as a landmark ruling. These attributes enable in-depth fairness audits, bias detection, and socio-legal research that go well beyond standard outcome classification or summarization tasks.

Institutional context—such as the specific High Court issuing the judgment—is annotated alongside demographic information like the accused’s gender and the type of bail sought (Regular, Anticipatory, or Interim). These features allow comparative analysis across courts, regions, and case types.

A complete overview of all annotated fields is provided in Table ~\ref{tab:schema}, which summarizes the schema, field types, and their legal relevance. This schema design facilitates a wide range of downstream applications across law, social science, and AI research.

\FloatBarrier


\section{Use Cases}

The \textit{IndianBailJudgments} dataset supports a wide spectrum of applications at the intersection of law, AI, and computational social science. Its multi-attribute design makes it suitable for both academic research and practical prototyping.

\subsection{Research Applications}

\begin{itemize}[leftmargin=1.5em]
    \item \textbf{Bail Outcome Prediction:} Supervised classification models can be trained to predict whether bail will be granted or rejected based on structured inputs such as crime type, prior record, gender, and court.

    \item \textbf{Legal Summarization:} Using fields like \texttt{facts}, \texttt{judgment\_reason}, and \texttt{summary}, both extractive and abstractive models can be trained for case-level summarization.

    \item \textbf{Bias and Fairness Analysis:} Fields like \texttt{bias\_flag}, \texttt{accused\_gender}, and \texttt{parity\_argument\_used} support socio-legal audits of systemic bias in bail jurisprudence.

    \item \textbf{Legal Argument Mining:} Researchers can analyze how legal issues are raised and how judgments are reasoned, enabling tasks such as entailment, contrastive reasoning, and argument similarity.

    \item \textbf{Information Extraction and Legal QA:} The dataset enables training of models that extract IPC sections, legal issues, and factual context from unstructured legal text, or answer fact-based questions from judgments.
\end{itemize}

\subsection{Educational Uses}

\begin{itemize}[leftmargin=1.5em]
    \item \textbf{Training for Law Students:} Law schools can use the dataset to teach bail jurisprudence, judicial rationale, and judicial writing patterns using real-world cases.

    \item \textbf{Legal NLP Curriculum:} AI/CS programs focusing on NLP can integrate the dataset into tasks like text classification, summarization, and bias detection with domain-specific complexity.

    \item \textbf{Legal Annotation Exercises:} The JSON structure can be used to guide legal annotation tasks in classrooms, helping students learn structured extraction from natural legal text.
\end{itemize}

\subsection{Prototype and Demonstration Systems}

\begin{itemize}[leftmargin=1.5em]
    \item \textbf{Interactive Legal Assistants:} The dataset can power LLM-based interfaces for answering bail-related legal queries or summarizing key case facts in plain English.

    \item \textbf{Explainable AI for Judges or Litigants:} Developers can build models that explain bail decisions based on prior reasoning patterns, supporting transparency and trust.

    \item \textbf{Document Pre-Filling Tools:} Based on input case facts, prototypes can auto-fill bail application templates using prior patterns and field values from the dataset.
\end{itemize}

This diversity of use cases makes \textit{IndianBailJudgments} a versatile resource for the legal-tech community, NLP researchers, and legal educators alike.

\section{Ethical Considerations}

While this dataset is derived from publicly available legal judgments, we acknowledge several important ethical considerations:

\begin{itemize}[leftmargin=1.5em]
    \item \textbf{Privacy:} Names and case details were sourced directly from public court records. Users must not repurpose the dataset for identifying or profiling individuals.
    
    \item \textbf{Non-Expert Annotation:} Although verified by legal personnel, the primary annotations were generated by a large language model and should not be considered legally authoritative.
    
    \item \textbf{Bias Propagation:} Judicial decisions may reflect inherent societal biases. The dataset preserves these patterns for study—not for deployment without critical analysis.
    
    \item \textbf{Responsible Usage:} This dataset is intended for academic research, educational use, and fair legal AI prototyping only. It must not be used for commercial, punitive, or real-world decision-making systems.
\end{itemize}

We urge all users to approach this dataset with legal awareness, sensitivity to potential harms, and a strong ethical commitment.

\section{Limitations}

Despite its contributions, the \textit{IndianBailJudgments} dataset comes with several limitations that must be acknowledged for responsible usage.

First, the dataset is limited to bail-related decisions from High Courts. While these decisions are typically well-reasoned and precedential, they exclude a significant volume of lower court and Sessions Court bail orders, which may reflect more day-to-day bail dynamics and procedural practices. This limits generalizability across the full spectrum of the Indian judiciary.

Second, while annotations were guided by a carefully designed prompt, fields such as \texttt{bias\_flag}, \texttt{landmark\_case}, and \texttt{legal\_issues} inherently involve interpretive judgment. Although consistency was enforced through prompt formatting and schema design, some degree of subjectivity or ambiguity remains—especially in borderline or complex cases.

Third, the dataset relies heavily on a large language model for annotation. While legal personnel validated a portion of the data, full manual expert verification was not performed at scale. This makes the annotations suitable for research and educational purposes, but not yet ready for deployment in real-world legal decision-making systems or legal advice tools.

Additionally, the dataset includes only English judgments, whereas Indian courts often use multilingual reasoning, especially in district or regional courts. This language limitation restricts linguistic diversity and leaves out large portions of India’s legal discourse.

\section{Future Work}

Several extensions are envisioned to enhance the scale, scope, and impact of the \textit{IndianBailJudgments} dataset.

First, the dataset will be extended to include bail decisions from Sessions and Metropolitan Magistrate Courts, capturing more routine legal patterns and reflecting ground realities across India's judicial landscape.

Second, multilingual versions of the dataset are planned—starting with Hindi, Marathi, and Bengali—by translating key judgment fields such as summaries and legal reasoning. This would make the dataset more inclusive and support research on cross-lingual legal understanding.

Third, plans are underway to incorporate richer legal reasoning features such as citation graphs, precedent mapping, and temporal case-linking. By encoding how prior judgments influence present decisions, we can enable deeper legal reasoning tasks such as precedent prediction and outcome counterfactuals.

Fourth, benchmark tasks are intended to be released, accompanied by baseline models for classification, summarization, bias detection, and information extraction. These tasks will support standardized evaluation and accelerate legal NLP research in India.

Lastly, we welcome community contributions—both legal and technical. Legal scholars can help refine annotations or define fairness evaluation metrics, while NLP researchers can explore architectural innovations, fine-tuning techniques, or low-resource learning paradigms.

\section{Conclusion}

This paper presents \textit{IndianBailJudgments}, the first structured dataset dedicated to Indian bail jurisprudence. By combining prompt-based large language model annotation with a legally grounded schema and partial human validation, we offer a high-utility resource for the legal AI and NLP communities.

The dataset’s rich annotations across over 20 fields—including legal reasoning, bail type, gender, crime category, and bias indicators—make it suitable for a wide range of tasks such as fairness analysis, summarization, judgment prediction, and legal education.

By building this dataset, the aim is to bridge the resource gap in Indian legal NLP, foster open research on judicial transparency, and support the responsible development of AI systems that can assist legal professionals, researchers, and public institutions alike.

We release this dataset openly, and encourage its use for academic, ethical, and constructive applications in the pursuit of justice, equity, and transparency in the Indian legal system.

\subsection*{Dataset Availability}

The full open-sourced dataset is available on Hugging Face and GitHub:

{\raggedright
\begin{itemize}[leftmargin=1.5em]
  \item \textbf{Hugging Face Dataset:} \url{https://huggingface.co/datasets/SnehaDeshmukh/IndianBailJudgments-1200}
  \item \textbf{GitHub Repository:} \url{https://github.com/SnehaDeshmukh28/IndianBailJudgments-1200}
\end{itemize}
}

We welcome community feedback and contributions.

\section*{Acknowledgments}

We thank the legal professionals who assisted in the manual review of annotations and contributed to the validation of schema design. We further acknowledge the broader open-source and research community for fostering tools and platforms that enabled the creation, hosting, and dissemination of this dataset. The dataset is publicly available to encourage further research in legal NLP.

\appendix
\section*{Appendix A: Annotation Prompt Used}

The following prompt was used to generate structured annotations from Indian bail-related judgments using GPT-4o. It was designed to handle both individual bail applications and principle-based landmark rulings, while maintaining consistency across more than 20 output fields in a valid JSON format.

\begin{quote}
You are assisting in the creation of a structured, high-quality dataset for Indian bail-related legal judgments intended for academic research and public release.

You will receive full or partial judgment text (sourced from Indian Kanoon or official court repositories). Your task is to extract structured information in strict JSON format using the following schema:

\textbf{Common Fields (all cases):}
\begin{itemize}
    \item \texttt{case\_id}: Leave blank (to be filled later)
    \item \texttt{case\_title}: Title of the case
    \item \texttt{court}: Court name (e.g., Bombay High Court)
    \item \texttt{date}: Date of judgment in YYYY-MM-DD
    \item \texttt{judge}: Judge’s name if mentioned; otherwise "Not Present"
    \item \texttt{ipc\_sections}: List of IPC/NDPS sections as strings (e.g., ["302", "498A"])
    \item \texttt{bail\_type}: One of "Regular", "Anticipatory", or "Interim"
    \item \texttt{bail\_cancellation\_case}: true if case involves review or cancellation of prior bail
    \item \texttt{landmark\_case}: true if the case discusses legal principles or sets precedent
\end{itemize}

\textbf{Case Type 1 – Fresh Bail Application:}
\begin{itemize}
    \item \texttt{accused\_name}: "Not specified" if unnamed
    \item \texttt{accused\_gender}: "Male", "Female", or "Unknown"
    \item \texttt{prior\_cases}: "Yes", "No", or "Unknown"
    \item \texttt{bail\_outcome}: "Granted" or "Rejected"

    (Note: For cancellation cases, set to "Rejected" if bail is cancelled, else "Granted")

    \item \texttt{bail\_outcome\_label\_detailed}: Free-text explanation (e.g., "Bail not cancelled", "Bail granted after FIR")
    \item \texttt{crime\_type}: One of ["Murder", "Sexual Offense", "Domestic Violence", "Narcotics", "Fraud or Cheating", "Attempt to Murder", "Theft or Robbery", "Dowry Harassment", "Kidnapping", "Extortion", "Cyber Crime", "Others"]
    \item \texttt{facts}: 3–4 sentence summary of the case background
    \item \texttt{legal\_issues}: Key legal questions (e.g., "Whether Section 437 prohibits bail in dowry cases")
    \item \texttt{judgment\_reason}: Legal reasoning behind the decision
    \item \texttt{summary}: Plain English 2-line summary of the judgment
    \item \texttt{bias\_flag}: true if caste, gender, or identity bias is observed
    \item \texttt{parity\_argument\_used}: true if co-accused parity was used in argument
\end{itemize}

\textbf{Case Type 2 – Landmark or Legal Principle Cases:}
\begin{itemize}
    \item \texttt{legal\_principles\_discussed}: List of doctrines (e.g., ["Anticipatory bail under Section 438 CrPC need not be time-bound"])
    \item \texttt{summary}: 2-line summary of what principle the judgment establishes
\end{itemize}

\textbf{Optional Fields:}
\begin{itemize}
    \item \texttt{region}: State or jurisdiction (e.g., "Maharashtra", "Uttar Pradesh")
\end{itemize}

\textbf{Output Formatting Guidelines:}
\begin{itemize}
    \item Include all fields, even if "Unknown", "Not applicable", false, or \texttt{[]}
    \item Use lowercase booleans (true / false)
    \item Use clean and valid JSON — do not include explanations or commentary
    \item Do not use null/NaN values; use appropriate placeholder strings
\end{itemize}

Return only the JSON object as the output.
\end{quote}

\bibliographystyle{unsrt}

\end{document}